# SEDA: A Self-Adapted Entity-Centric Data Augmentation for Boosting Gird-based Discontinuous NER Models

Wen-Fang Su[1,*], Hsiao-Wei Chou[2,*], and Wen-Yang Lin[3]

## Abstract

Named Entity Recognition (NER) is a critical task in natural language processing, yet it remains particularly challenging for discontinuous entities. The primary difficulty lies in text segmentation, as traditional methods often missegment or entirely miss cross-sentence discontinuous entities, significantly affecting recognition accuracy. Therefore, we aim to address the segmentation and omission issues associated with such entities. Recent studies have shown that grid-tagging methods are effective for information extraction due to their flexible tagging schemes and robust architectures. Building on this, we integrate image data augmentation techniques—such as cropping, scaling, and padding—into grid-based models to enhance their ability to recognize discontinuous entities and handle segmentation challenges. Experimental results demonstrate that traditional segmentation methods often fail to capture cross-sentence discontinuous entities, leading to decreased performance. In contrast, our augmented grid models achieve notable improvements. Evaluations on the CADEC, ShARe13, and ShARe14 datasets show F1 score gains of 1–2.5% overall and 3.7–8.4% for discontinuous entities, confirming the effectiveness of our approach.

## 1 Introduction

Named entity recognition (NER) is a crucial task in the field of natural language processing, aiming to locate and classify named entities into predefined categories from text. In recent years, NER research has been subdivided into various task types, including flat [13, 24], overlapping [23, 32], and discontinuous [5, 14] NER tasks, with discontinuous NER seen as the most challenging among them. As shown in Figure 1, entities in the sentence are discontinuous; their representation might be nested, overlapping, or even span multiple sentences. This diversity significantly increases the difficulty of the recognition task.

E2

A patient at the downtown health clinic reports severe muscle \n

E1

pain in their legs and ankles.

**Figure 1: Example showing two discontinuous entities.**

Several studies have adopted the preprocessing script proposed by Dai et al. [5] to segment datasets, employing specific character retention and sentence tokenization strategies [14, 15, 30, 31, 33]. Typically, a simple newline character is used as the sentence delimiter. For example, in Figure 1, the sentence is divided into "Sentence one: A patient at the downtown health clinic reports severe muscle" and "Sentence two: pain in their legs and ankles.", which results in the separation of two discontinuous entities. Consequently, when cross-sentence discontinuous entities occur, this method fails to correctly identify the entity, thereby affecting the model's predictions. Therefore, preprocessing methods for discontinuous entities are particularly tricky, requiring special consideration to maintain entity integrity after text processing. A literature review reveals that most studies focus on enhancing model architectures or developing related auxiliary loss functions [7, 11, 20, 23, 25, 26, 28, 31, 32], with few discussing data processing methods.

Building on innovative model architectures like the Unified Word-Word Framework (Word2NER) [15], which reformulates discontinuous named entity recognition (NER) by modeling inter-word relationships through a grid structure, we explore graph-based preprocessing for discontinuous entity recognition. Word2NER's graph-inspired representation and use of CNN techniques have demonstrated strong NER performance. Inspired by this, we propose a preprocessing method that employs image augmentation techniques, such as cropping and scaling within the grid, combined with a self-learning strategy to enhance model performance and improve the recognition of discontinuous entities [2, 21].

Figure 2(a) illustrates the image recognition of a cat, where the target object is enclosed within the red box, and the blue area represents peripheral vision. Incorporating peripheral vision into the field of computer vision (CV) has demonstrated significant improvements in recognition performance [19, 22]. This concept is extended to grid-tagging models, as demonstrated in Figure 2(b). In the sentence, "I do experience stomach pain from time to time," the entity "stomach pain" is mapped onto a grid. The red area highlights the critical region, while the surrounding blue blocks act as supplementary intervals, extending the focus area in a manner analogous to peripheral vision. Other areas are treated as background information.

* Both authors contributed to this work equally.
1 Dept. of Computer Science and Information Engineering, National University of Kaohsiung, Kaohsiung City, Taiwan
email: fangfangsusu@gmail.com
2 Dept. of Computer Science and Information Engineering, National Taiwan University of Science and Technology, Taipei, Taiwan
email: victor88041559@gmail.com
3 Dept. of Computer Science and Information Engineering, National University of Kaohsiung, Kaohsiung City, Taiwan
email: wylin@nuk.edu.tw

† The programs and data presented in this paper are openly available at https://github.com/fang1204/SEDA

In summary, our research contributions can be highlighted in three key points: First, to the best of our knowledge, we are the first to propose the application of image augmentation methods for discontinuous entity recognition. Second, our approach leverages a variety of image augmentation techniques to overcome the limitations of grid models in discontinuous entity recognition. Finally, we demonstrate the generalization and effectiveness of our method across diverse datasets and grid models.

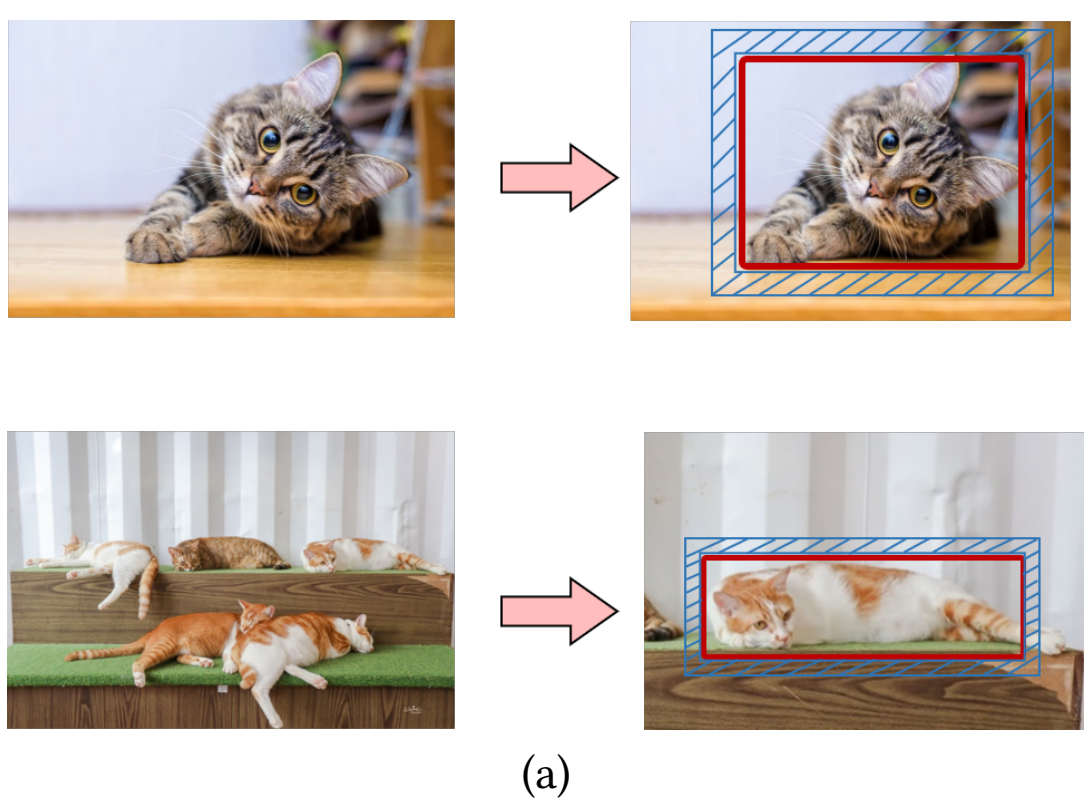

(a)

I do experience stomach pain from time to time.

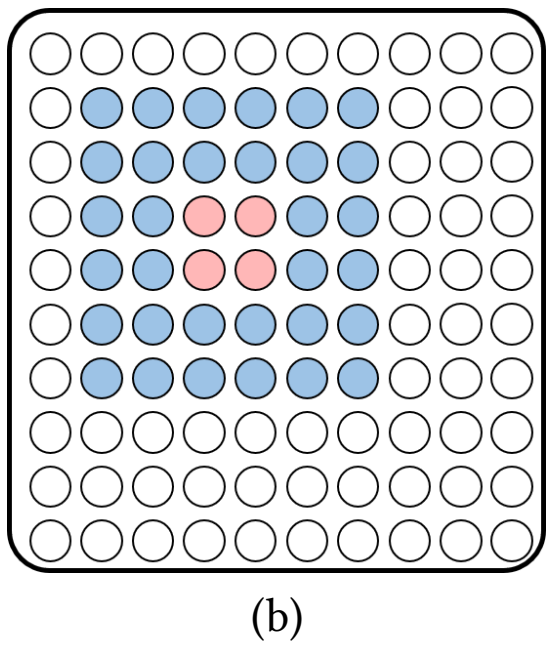

(b)

**Figure 2: The object boundaries in the image correspond to the entity boundaries in the grid.**

## 2 Grid-Tagging Method and Related work

The goal of the grid model is to create a function $f$ that, given an input document $D$, outputs predicted entities $E$:

$$E \leftarrow f(D) \tag{1}$$

It consists of the following three structures:

**(1) Encoder Layer** First, the text is input into a pre-trained encoder model. The input sentence is represented as $X = \{x_1, x_2, ..., x_N\}$, where $x_i$ represents each token or word piece. After processing through the model, these $x_i$ are transformed into a series of vectors $H = \{h_1, h_2, ..., h_N\} \in R^{N \times d_h}$ through max pooling and bidirectional long short-term memory network (Bi-LSTM). Here, $d_h$ represents the dimension of the word vector.

**(2) Convolution Layer** The process then enters a convolutional neural architecture. It first uses the Conditional Normalization Layer (CLN) proposed by [18] to construct relationships between words. A matrix representation $V \in R^{N \times N \times d_h}$ is defined, where $V_{ij}$ represents the vector representation of the relationship between the $i$-th word and the $j$-th word. The following formula is defined:

$$V_{ij} = \mathrm{CLN}(h_i, h_j) = \gamma_{ij} \odot (\frac{h_j - \mu}{\sigma}) + \lambda_{ij} \tag{2}$$

where $h_i$ is the representation of word $x_i$, $h_j$ is the representation of word $x_j$, and $\gamma_{ij}$ and $\lambda_{ij}$ are the gain parameter and bias for normalization, which are jointly determined by $h_i$ and $h_j$. $\mu$ and $\sigma$ are the mean and standard deviation across the elements of $h_j$. Position embedding $E^d$ and region embedding $E^t$ are added, where $E^d \in R^{N \times N \times d_{ed}}$ and $E^t \in R^{N \times N \times d_{et}}$. These three embeddings are then concatenated and passed through a multi-layer perceptron (MLP) to reduce their dimensionality, obtaining a position-region-aware representation of the grid $C \in R^{N \times N \times d_c}$, as per the following formula:

$$C = \mathrm{MLP}_1([V; E^d; E^t]) \tag{3}$$

Subsequently, multiple dilated convolutions[12] are used to compute $C$:

$$Q_l = \sigma(\mathrm{DConv}_l(C)) \tag{4}$$

where $l$ represents different dilation factors, typically $[1, 2, 3]$, and $\sigma$ is the GELU activation function [9]. Finally, the ultimate word pair grid representation $Q = [Q_1, Q_2, Q_3] \in R^{N \times N \times 3d_c}$ is obtained.

**(3) Co-Predictor and Learning** After establishing the grid vector, the obtained features are input again into a multi-layer perceptron (MLP) and a bilinear predictor to enhance relationship classification. The following formulas are defined:

$$y'_{ij} = \mathrm{MLP}_2(Q_{ij}) \tag{5}$$

$$y''_{ij} = s_i^T U o_j + W[s_i; o_j] + b \tag{6}$$

where $U$, $W$, and $b$ are trainable parameters, and $s_i = \mathrm{MLP}_3(h_i)$ and $o_j = \mathrm{MLP}_4(h_j)$ represent the subject and object representations, respectively. Combining the scores from the MLP and bilinear predictor yields the final prediction probability:

$$y_{ij} = \mathrm{Softmax}(y'_{ij} + y''_{ij}) \tag{7}$$

Grid-tagging method can be traced back to Wang et al. [29], which transformed entity boundaries into a token pair linking problem. The following year, Wang et al. [30] further proposed using two grids to predict entity boundaries and entity word relationships separately, and decoding complete entities from entity fragment graphs through maximal clique discovery. Later, Li et al. [15] also based their work on grid-tagging and further transformed discontinuous named entity recognition (NER) into a word-to-word relationship identification problem, setting up two word relationships: Next-Neighboring-Word (NNW) and Tail-Head-Word (THW), using a single grid to include all word-to-word relationships. Finally, Liu et al. [17] expanded on Li et al.'s word-to-word relationships by adding two more labels: Previous-Neighboring-Word (PNW) and Head-Tail-Word (HTW), to enrich the representation of inter-word relationships.

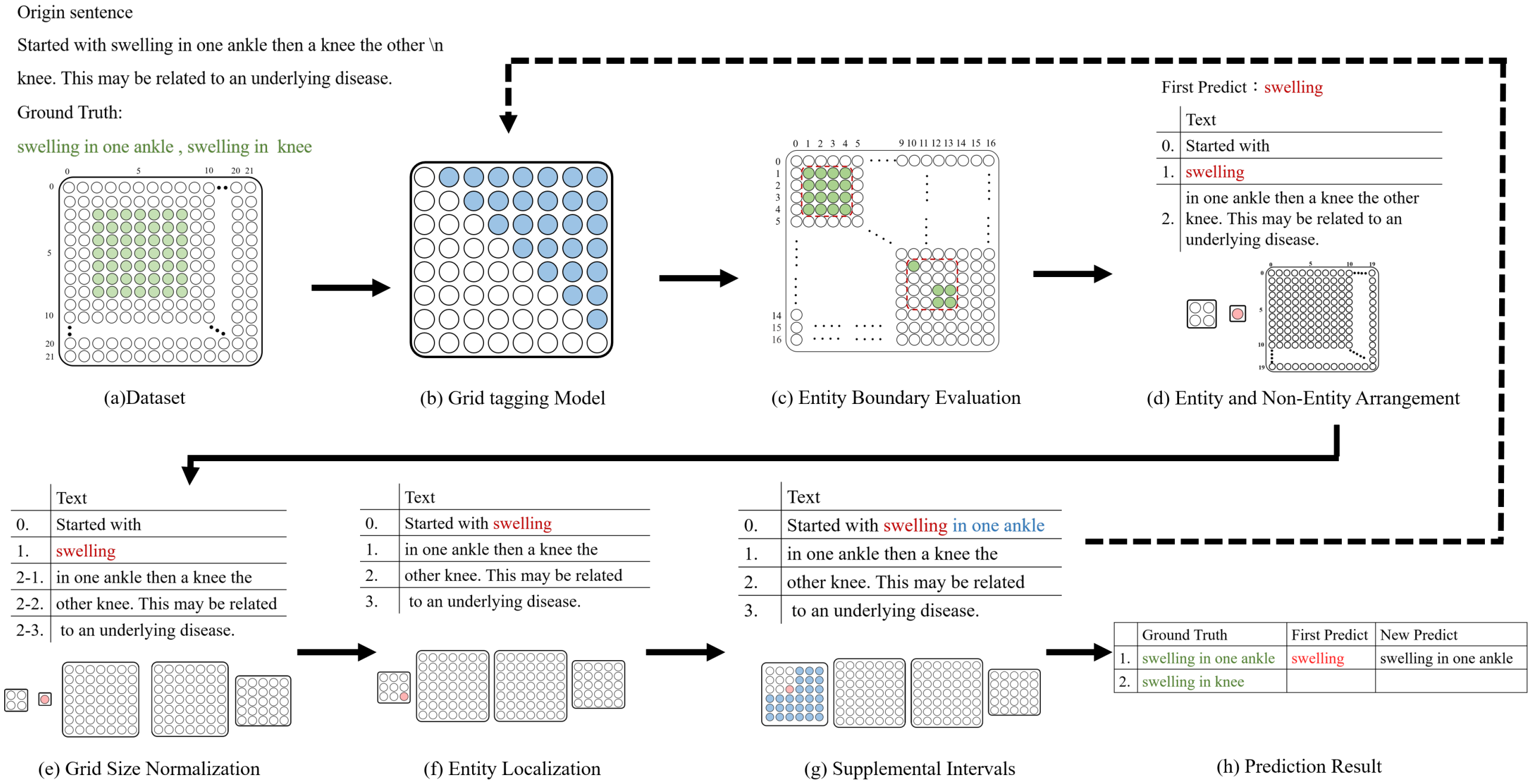

**Figure 3: Flowchart of the method.**

## 3 Methodology

Next, we introduce how image data augmentation techniques are integrated with the grid-based model, and how we effectively capture missing discontinuous entities. This approach gradually trains the model to be more focused on discontinuous entities, and is therefore termed **SEDA**. Figure 3 illustrates the overall workflow of the method. First, the grid labeling model generates initial target entities. Then, a relatively lenient calculation method is applied to evaluate the quality of the entity boundaries. The results are subsequently sorted by odd and even positions: even positions correspond to non-entity sentences, while odd positions indicate target entities. This sorting facilitates grid size normalization, entity localization adjustments, and corrections for non-entity sentences. The specific procedures will be detailed in the following sections. Additionally, we iteratively optimize the prediction results by using the outcome of each enhancement step as a new baseline, and repeatedly applying the process described in Sections 3.1 to 3.4 to further improve model performance.

### 3.1 Entity Boundary Evaluation

Firstly, as shown in Figures 3(a) to (c), we use a grid-tagging model to generate initial target entities and obtain better entity boundary positions through entity boundary evaluation. To evaluate these results, we designed a unique set of scoring metrics—Entity Boundary F1 score (EBF), Entity Boundary Precision (EBP), and Entity Boundary Recall (EBR), where EBF is the primary scoring metric used in the initial boundary prediction phase as well as in the subsequent Section 3.4. Unlike the traditional F1 score, EBF employs a more lenient strategy, allowing for the prediction of more potential entities. Specifically, we only consider whether the head and tail tokens of the predicted entity match those of the correct entity.

Assuming there are $|\mathcal{D}|$ number of documents, each with correct entities $E = \{e_1, ..., e_n\}$, where $n$ is the number of answer entities, and predicted entities $E' = \{e'_1, ..., e'_m\}$, where $m$ is the number of predicted answer entities. Extract the last word of each entity from texts $E$ and $E'$, resulting in $G = \{g_1, ..., g_n\}$, the last words of the correct entities, and their corresponding last words of predicted entities $P = \{p_1, ..., p_m\}$, with a specific formula design as follows:

$$EBP = \frac{1}{|\mathcal{D}|m} \sum_{d}^{|\mathcal{D}|} \sum_{\alpha}^{n} \sum_{\beta}^{m} S(p_\alpha, g_\beta) \tag{8}$$

$$EBR = \frac{1}{|\mathcal{D}|n} \sum_{d}^{|\mathcal{D}|} \sum_{\alpha}^{n} \sum_{\beta}^{m} S(p_\alpha, g_\beta) \tag{9}$$

$$S(p_\alpha, g_\beta) = \begin{cases} 1 & p_\alpha = g_\beta \\ 0 & p_\alpha \neq g_\beta \end{cases} \tag{10}$$

$$EBF = \frac{2 \times EBP \times EBR}{EBP + EBR} \tag{11}$$

### 3.2 Grid Size Normalization

The purpose of grid size normalization is to constrain grid dimensions within a specific range. Previous studies have found that image sizes in datasets can significantly influence model training, where larger images may lead to background interference and feature redundancy, subsequently affecting the model's predictive performance [16]. Inspired by these findings, our approach first arranges the predicted entity boundaries in an odd-even pattern: sentences in odd-numbered positions store the identified target entities , while those in even-numbered positions contain the remaining text, which tends to be longer. This arrangement preserves the integrity of complete entity segments and avoids cutting through target entities when trimming grids. As shown in Figure 3(d), we extract these predicted entities as a basis for segmenting the text. For grid size constraints, we split sentences in even-numbered positions into blocks according to varying text lengths, refer to Appendix B. For example, the standard size for a sample text is 7. In the sentence "In one ankle then a knee the other knee. This may be related to an underlying disease," which exceeds the standard size, the text is divided into three segments: 2-1, 2-2, and 2-3, as illustrated in Figure 3(e). This method effectively maintains grid size within the specified range, enhancing both training and prediction efficiency.

Table 1: Complete statistics of three datasets.

| | CADEC | | | | ShARe13 | | | | ShARe14 | | | |
|---|---|---|---|---|---|---|---|---|---|---|---|---|
| | All | Train | Dev | Test | All | Train | Dev | Test | All | Train | Dev | Test |
| #Entities | 6,318 | 4,430 | 898 | 990 | 11,161 | 5,146 | 675 | 5,340 | 19,157 | 10,354 | 810 | 7,993 |
| #Discontinuous | 675 | 489 | 93 | 93 | 1,090 | 560 | 93 | 437 | 1,710 | 992 | 93 | 635 |
| -cross-sentence | 0 | 0 | 0 | 0 | 12 | 5 | 0 | 7 | 83 | 12 | 0 | 71 |
| %Discontinuous | 10.7 | 11 | 10.4 | 9.4 | 9.8 | 10.9 | 13.8 | 8.2 | 8.9 | 9.6 | 11.5 | 7.8 |

## 3.3 Entity Localization

Next, we design the placement of predicted entities at predetermined locations, primarily positioning the predicted entities later, detailed location choices can be seen in Appendix A. The purpose of this stage is to allow the model to learn the regularity of the data, enhancing its training efficiency. We merge the odd and even-positioned sentences, arranging them in the order of even sentences first, followed by odd sentences, ensuring that each predicted entity is positioned at the end of the sentence, as the change from (e) to (f) in Figure 3 shows, sequentially connecting even sentence "0" to odd sentence "1". For cases like "2-1", "2-2", "2-3" where the sentences are even-numbered and do not contain predicted entities, and no predicted entities are subsequently found, no action is taken in this stage. Simultaneously, we also define two types of sentences: ES (sentences containing predicted entities) and NES (sentences not containing predicted entities), for subsequent use.

## 3.4 Supplemental Intervals

Subsequently, to prevent the unintended truncation of entities during text segmentation and to ensure the integrity of the predicted entities, we draw inspiration from the concept of magnification in image data augmentation to adjust grid size, as shown in (g) of Figure 3. The blue blocks in the example represent the supplemental intervals. Specifically, we designed strategies for pre-supplemental and post-supplemental intervals. For sentences containing predicted entities (ES), one can choose to apply either a pre-supplemental or a post-supplemental; for sentences that do not contain predicted entities (NES), it's possible to apply either or both types of supplements. In the figure, the setting is to apply a post-supplemental of 3 to ES. This method effectively compensates for the parts of predicted entities that might be missed or lacking.

Lastly, further optimization is achieved iteratively, as indicated by dashed lines in the diagram, where the current prediction is refined by intersecting it with the previous one. This gradual refinement aims to achieve the correct answer. The upcoming experimental analysis will compare the results of single enhancement (SEDA-Once) and multiple enhancements (SEDA-Mul).

# 4 Experiments and Results Analysis

## 4.1 Dataset Introduction

To validate the effectiveness of our method, we conducted experiments on three Named Entity Recognition (NER) datasets from the biomedical and clinical domains. These datasets—CADEC, Share/CLEF 2013 (ShARe13), and Share/CLEF 2014 (ShARe14)—feature discontinuous, nested, and flat entities. This study builds upon previous research on Adverse Drug Events (ADEs) and Disorder entities. Table 1 presents the number of entities in each dataset and the proportion of discontinuous entities. We have corrected the test data counts for previous datasets, adjusting the number for **ShARe13 from 5,333 to 5,340, and for ShARe14 from 7,922 to 7,993**. This differs from the statistics reported in Dai et al. [5].

## 4.2 Is Semantic Nature Important?

In many NER approaches, much emphasis is placed on the semantic properties of entities. However, for **discontinuous NER tasks**, an over-reliance on semantics might overlook the structural importance of correctly identifying entity boundaries. Our approach shifts the focus away from the assumption that semantic information is essential by introducing the Entity Boundary F1 score (EBF). EBF emphasizes the head and tail positions of entities, regardless of their semantic attributes, ensuring that models concentrate on capturing accurate entity boundaries rather than relying solely on semantic cues.

This boundary-centric view allows the model to improve its recognition of discontinuous entities, especially when the semantic meaning is less clear or entities span multiple sentences. As a result, EBF becomes a crucial metric in evaluating the precision and recall of entity boundaries in our model, providing a more lenient yet effective strategy for discontinuous entity recognition. In the following experiments, we will demonstrate how EBF contributes to better performance in recognizing discontinuous entities, proving that focusing on entity boundaries is a more effective approach than relying solely on semantic information.

**Table 2: The table shows the results of different datasets under two grid architectures, with bold numbers representing the highest scores in each column. "*" denotes significance at $p - value < 0.05$ and "**" denotes < 0.01 compared to the original grid model. Our scores presented are based on calculations from all test entities. "†" denotes our reproduced scores.**

| METHOD | CADEC | | | ShARe13 | | | ShARe14 | | |
|---|---|---|---|---|---|---|---|---|---|
| | P | R | F1 | P | R | F1 | P | R | F1 |
| TOE† | 77.47 | 70.51 | 73.82 | 82.04 | 80.52 | 81.27 | 78.38 | 82.52 | 80.39 |
| + SEDA-Once | 76.48 | 71.92 | 74.13 | 82.57 | 81.74 | *82.15 | **80.34** | 80.91 | 80.62 |
| + SEDA-Mul | 77.07 | 71.62 | *74.24 | **86.62** | 79.05 | 82.67 | 79.46 | 83.26 | *81.32 |
| W2NER† | 76.06 | 70.91 | 73.39 | 81.91 | 81.16 | 81.53 | 79.38 | 81.63 | 80.49 |
| +Label-wise token rep. | 77.14 | 69.19 | 72.95 | 83.99 | 80.59 | 82.26 | 78.77 | 82.43 | 80.56 |
| +Synonym replacement | 74.89 | 72.02 | 73.43 | 84.33 | 79.34 | 81.76 | 77.62 | 83.44 | 80.42 |
| +Mention replacement | 75.90 | 70.61 | 73.16 | 84.27 | 79.64 | 81.89 | 78.30 | 83.41 | 80.77 |
| +Shuffle within segments | 76.38 | 71.52 | 73.87 | 84.24 | 79.69 | 81.90 | 78.54 | 82.35 | 80.40 |
| +DAGA | 74.10 | 70.81 | 72.42 | 83.92 | 77.79 | 80.74 | 78.74 | 83.24 | 80.93 |
| +MELM | **77.57** | 68.48 | 72.75 | 83.95 | 79.53 | 81.68 | 79.22 | 83.21 | 81.16 |
| +SEDA-Once | 75.95 | **72.72** | *74.15 | 82.73 | 82.45 | *82.59 | 78.46 | **83.81** | *81.05 |
| +SEDA-Mul | 75.84 | 72.93 | ****74.36** | 82.92 | **82.67** | ****82.80** | 80.30 | 80.90 | ****81.61** |

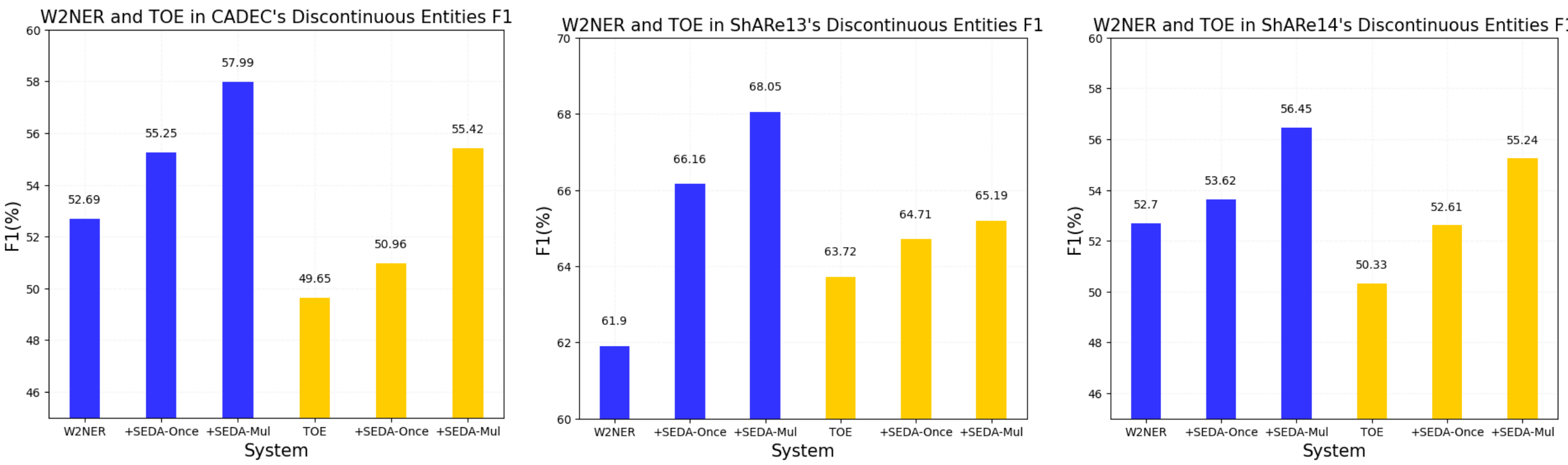

**Figure 4: The results for discontinuous entities.**

### 4.3 Backbone Models & Common Augmentation Methods

We apply our method to two state-of-the-art (SOTA) grid models commonly used in discontinuous NER and demonstrate its effectiveness through experimental results.

**Backbone Models:**

(1) **W2NER** [15]: TThis model represents adjacency relationships between entity words as a two-dimensional grid and refines the grid representation through multi-granularity two-dimensional convolution operations to capture complex relationships between entities.

(2) **TOE** [17]: Based on the W2NER model, this model constructs additional textual relations and incorporates a Tag Representation Embedding Module (TREM) to enhance the model's understanding and representation of entity relationships.

**Common Augmentation Methods:**

(1) **Label-wise Token Replacement**: Replaces specific tokens based on labels to generate similar but semantically distinct sentences, enhancing the model's ability to distinguish between labels [4].

(2) **Synonym Replacement**: Replaces words in the original sentence with synonyms to maintain meaning while introducing diversity [4].

(3) **Mention Replacement** Replaces entities mentioned in the text with other entities of the same category to improve the model's generalization capability [4].

(4) **Shuffle Within Segments** Randomly shuffles word order within segments of a sentence, preserving semantics while enhancing the model's robustness to word order [4].

(5) **DAGA (Data Augmentation with Generative Adversarial Network)** Uses generative adversarial networks to generate text similar to the original sentence, increasing data diversity [6].

(6) **MELM (Masked Entity Language Model)** Generates new sentences by masking entities, preserving essential information while enhancing variation in the corpus [34].

## 4.4 Results and Analysis

The main results[1] of various augmentation methods applied to two backbone models are shown in Table 2. Furthermore, to verify the accuracy of discontinuous entities under our approach, we extracted the discontinuous entities from three datasets and evaluated the improvements in performance, as shown in Figure 4. Cross-analyzing the results of these experiments, we can observe that our method achieves the best F1 scores across the three datasets, along with consistent improvements in discontinuous entity recognition. This indicates that our approach not only enhances predictions through image data augmentations but also improves discontinuous entity prediction through techniques such as grid cropping, fixed entity positions, and supplementary intervals to capture more potential discontinuous entities, leading to higher prediction accuracy. The consistent results across all three datasets demonstrate the generalizability of our method. As a result, the performance of W2NER's F1 scores improved by 1.79%, 1.06%, and 0.56% on CADEC, ShARe13, and ShARe14, respectively, while TOE improved by 0.43%, 0.88%, and 0.23%. After multiple enhancements, W2NER's improvements reached 2.48%, 1.27%, and 1.12%, and TOE's improvements reached 0.95%, 1.4%, and 0.93%. This demonstrates that our technique can enhance prediction accuracy by approximately 1%.

## 4.5 Comparison of Cross-Sentence Discontinuous Entities

Figure 5 illustrates the average coverage and accuracy rates for cross-sentence discontinuous entities in the ShARe13 and ShARe14 datasets. To evaluate our method, we compared it with the mainstream data processing approach proposed by Dai et al. [5]. For this analysis, we separately calculated the coverage and prediction accuracy for these entities. It is evident that the method in Dai et al. [5] failed to predict such entities due to sentence splitting caused by newline characters during preprocessing, resulting in 0% coverage and accuracy. In contrast, significant improvements in both metrics were observed when using SEDA-Once and SEDA-Mul.

## 4.6 Brief

Returning to the previously mentioned question, "Is Semantic Nature Important in Discontinuous Entity?" our experiments have shown varied results. Specifically, when our method employed graph augmentation on the data, the semantic expression of sentences was indeed altered. However, the experimental results indicate that discontinuous entities consistently improved after multiple iterations. This demonstrates that the connection between the existence of discontinuous entities and semantics is not strong, and further validates the effectiveness of our approach in identifying discontinuous entities.

## 4.7 Impact of Supplemental Intervals

Finally, in Table 3, we conducted experiments with different supplemental interval settings as well as experiments without using EBF, using the scores under SEDA-Once as the benchmark, focusing on W2NER as the research subject. We applied different settings to ES, NES, and BOTH (ES and NES), including "only look forward" (unidirectional forward supplementation), "only look backward" (unidirectional backward supplementation), and "look both sides" (bidirectional supplementation), examining the effects of varying interval sizes. The results show that "look both sides" consistently outperformed "only look forward" and "only look backward." In the CADEC dataset, the BOTH setting with "look both sides" achieved the best results, while in the ShARe13 and ShARe14 datasets, the ES setting with "look both sides" had the highest F1 score. Additionally, we also demonstrated the impact of different interval sizes under the "look both sides" (BOTH) setting. The best parameter results are shown in Table 8. Based on the identified optimal parameters, we further experimented with the model enhancement process without using EBF [2]. We observed that the overall scores for all three datasets decreased, which demonstrates the effectiveness of using EBF in the enhancement process.

[1] For a comparison with previous scores, please refer to Appendix C.

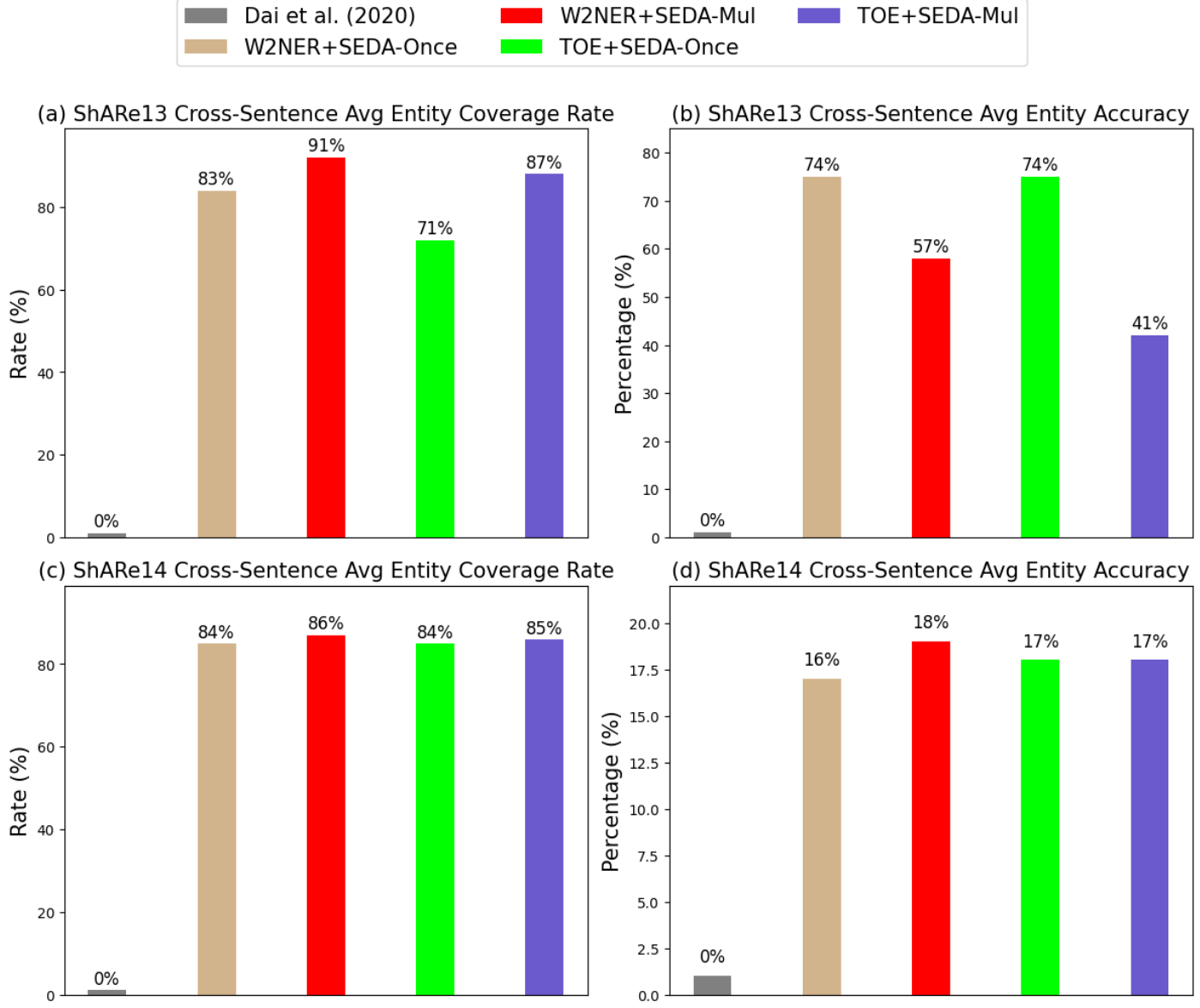

**Figure 5: Average coverage and accuracy rates for cross-sentence discontinuous entities in ShARe13 and ShARe14 using our method.**

## 5 Conclusion

In this paper, we propose a novel approach that applies the concept of image augmentation techniques to grid-tagging models for NER, which has not been previously introduced in earlier works. This method results in more comprehensive prediction outcomes. After conducting evaluations and analyses across three different datasets, it was found that the integration of enhancement techniques significantly improved the prediction results. Ablation experiments further validated the effectiveness of these enhancements. Further experimental analysis indicates that our proposed model is better at identifying discontinuous entities.

[2] Adopting the highest F1 score on the validation set instead of the highest EBF in the enhancement process.

**Table 3: Comparison table of scores for ablative experiments on the W2NER model. The scores in parentheses represent the corresponding validation set scores.**

| | CADEC | | | ShARe13 | | | ShARe14 | | |
|---|---|---|---|---|---|---|---|---|---|
| | **P** | **R** | **F1** | **P** | **R** | **F1** | **P** | **R** | **F1** |
| Paper Setting | 75.95 | **72.72** | **74.15** | 82.73 | **82.45** | **82.59** | 78.46 | 83.81 | **81.05** |
| | (72.46) | (70.60) | **(71.52)** | **(81.25)** | (79.67) | (80.45) | (81.38) | (80.05) | **(80.71)** |
| w/o EBF | 70.45 | 73.43 | 71.91 | 82.74 | 81.42 | 82.07 | 77.52 | 84.11 | 80.64 |
| | (68.24) | (71.50) | (69.61) | (78.01) | (79.52) | (78.76) | (80.62) | (78.29) | (79.44) |
| only look forward =2 (ES) | 71.07 | 73.94 | 72.48 | 82.51 | 81.85 | 82.18 | 79.07 | 82.27 | 80.64 |
| | (71.33) | (69.82) | (70.57) | (79.49) | (78.77) | (79.13) | (84.57) | (75.66) | (79.87) |
| only look forward =2 (NES) | **73.71** | 70.81 | 72.23 | 82.84 | 81.04 | 81.93 | 77.48 | 83.87 | 80.55 |
| | (68.87) | (70.71) | (69.78) | (77.75) | (79.37) | (78.55) | (79.62) | (78.92) | (79.27) |
| only look backward =2 (ES) | 70.81 | 74.75 | 72.73 | 82.40 | 82.24 | 82.32 | 77.03 | **85.07** | 80.85 |
| | (70.33) | (71.27) | (70.80) | (78.43) | (80.42) | (79.41) | (80.82) | (79.30) | (80.05) |
| only look backward =2 (NES) | 71.37 | 74.04 | 72.68 | **83.57** | 80.95 | 82.24 | 77.03 | 84.98 | 80.81 |
| | (69.15) | (72.38) | (70.73) | (79.58) | (79.22) | (79.40) | (81.06) | (78.92) | (79.97) |
| look both side =2 (ES) | 70.88 | 74.75 | **72.76** | 82.73 | **82.45** | **82.59** | 78.46 | 83.81 | **81.05** |
| | (69.49) | (72.27) | (70.85) | **(81.25)** | (79.67) | **(80.45)** | (81.38) | (80.05) | **(80.71)** |
| look both side =2 (NES) | 71.00 | 73.94 | 72.44 | 82.45 | 82.21 | 82.33 | **78.02** | 83.76 | 80.79 |
| | (69.46) | (71.16) | (70.30) | (79.79) | (79.07) | (79.43) | (82.49) | (77.42) | (79.87) |
| look both side =2 (BOTH) | 72.04 | 74.44 | 73.22 | 82.73 | 82.26 | 82.49 | 79.46 | 82.61 | 81.00 |
| | (68.67) | (73.72) | (71.11) | (81.75) | (78.67) | (80.18) | (80.60) | (80.30) | (80.45) |
| look both side =3 (BOTH) | 74.95 | 72.83 | 73.87 | 82.76 | 82.11 | 82.44 | **79.99** | 81.92 | 80.94 |
| | (69.72) | (72.83) | (71.24) | (78.67) | (81.61) | (80.12) | (80.15) | **(80.55)** | (80.35) |
| look both side =4 (BOTH) | 75.95 | **72.72** | **74.15** | 83.18 | 81.59 | **82.37** | 78.43 | 83.36 | 80.82 |
| | **(72.46)** | (70.60) | **(71.52)** | (80.24) | (78.92) | (79.58) | (80.72) | (79.30) | (80.00) |

# A Impact of Entity Boundaries

**Table 4: The influence of CADEC boundaries on entities.**

| | P | R | F1 |
|---|---|---|---|
| Origin | 71.01 | 74.24 | 72.59 |
| Masking before the first entity | 76.71 | 72.53 | 74.56 |
| Masking after the last entity | 78.92 | 73.74 | 76.15 |
| Masking on both sides | 78.99 | 79.39 | 79.19 |

At the beginning of our research, we explored the impact of entity boundaries on model predictions to enhance the efficiency of the model in learning key features. We designed experiments using the CADEC dataset, where parts of the given text unrelated to entities were masked. The experiments included masking the content before the first entity in the text, after the last entity, and both ends simultaneously, to observe the impact of boundaries on model predictions. The results, as shown in Table 4, indicate that each masking effect significantly influences entity prediction. It is observed that the model performs best when both ends are masked, with an F1 score of 0.7919; followed by masking only after the last entity with a score of 0.7615, and finally masking only before the first entity with a score of 0.7456. These findings motivate us to further explore whether it can enhance model performance if we choose a relatively ample approach in predictions by positioning the prediction of entity locations posteriorly, simulating the situation through boundary confirmation by answers.

# B Grid Size Setting

**Table 5: Setting grid sizes corresponding to document sizes.**

| Document length | Grid Size |
|---|---|
| ~200 | 7 |
| 200~350 | 9 |
| 350~500 | 11 |
| 500~1000 | 13 |
| 1000~1350 | 15 |
| 1350~1500 | 16 |
| 1500~2000 | 17 |
| 2000~ | 19 |

Table 5 presents the appropriate grid size settings for different document lengths. To normalize grid sizes across various document length ranges, we imposed constraints based on statistical analysis and sampling conducted across three datasets. Detailed parameter settings are provided in Table 6. For instance, we randomly selected three texts (A, B, and C) with lengths between 0 and 50, trained and predicted them using different grid sizes (e.g., 7, 8, and 9), and recorded their convergence times. The grid size with the shortest convergence time in most cases was selected as the optimal choice. For example, texts A and B show the shortest convergence times at a grid size of 7, so for texts with lengths between 0 and 50, we set the grid size to 7.

**Table 6: Grid size constraint perusing experiments concerning different document lengths.**

| Document length (0∼50) | Grid Size | convergence times | lowest convergence times |
|---|---|---|---|
| A | 7 | 5 | ✔ |
| | 8 | 10 | |
| | 9 | 15 | |
| B | 7 | 6 | ✔ |
| | 8 | 11 | |
| | 9 | 16 | |
| C | 7 | 10 | |
| | 8 | 9 | ✔ |
| | 9 | 17 | |

## C Scores Under Unified Evaluation

**Table 7: Scores Under Unified Evaluation: This table presents the scores calculated using the unified entity statistics from Table 11 in Dai et al. [5], alongside those reported by previous studies that adopted the same evaluation criteria. For CADEC, the number of test entities remains unchanged, while Share13 and Share14 use 5,333 and 7,922 test entities, respectively. As discussed in Section 4.5, some cross-sentence discontinuous entities are excluded in earlier studies to facilitate consistent statistics and score computation. Accordingly, we recomputed the scores of our method under this unified evaluation protocol and present the results here.**

| METHOD | CADEC | ShARe13 | ShARe14 |
|---|---|---|---|
| | F1 | F1 | F1 |
| Hu et al. [10] | 73.75 | 83.02 | 82.20 |
| Wang and Lu [27] | 58.00 | 70.30 | 74.70 |
| Yan et al. [31] | 70.64 | 79.69 | 80.34 |
| Fei et al. [7] | 72.40 | 80.30 | — |
| Li et al. [14] | 69.90 | 82.50 | — |
| He and Tang [8] | 73.15 | 80.63 | — |
| Corro [3] | 72.9 | 82.1 | 80.9 |
| Cabral et al. [1] | 73.43 | **83.22** | **82.54** |
| TOE† | 73.82 | 81.39 | 80.49 |
| + SEDA-Once | 74.13 | 82.22 | 80.77 |
| + SEDA-Mul | 74.24 | 82.78 | 81.51 |
| W2NER† | 73.39 | 81.71 | 80.75 |
| +SEDA-Once | 74.15 | 82.76 | 81.21 |
| +SEDA-Mul | **74.36** | 82.91 | 81.84 |

## D Parameter & Model Settings

In terms of experimental settings, the detailed configuration of parameters is shown in Table 8. In the table, the ES/NES settings of 1 and 0 indicate whether supplemental intervals were used, with 1 representing the use of supplemental intervals and 0 indicating that they were not used. "Look forward" and "look backward" correspond to the number of intervals for forward and backward supplementation, respectively. For additional model parameters, please refer to Table 9.

**Table 8: Parameter settings.**

| | CADEC | ShARe13 | ShARe14 |
|---|---|---|---|
| ES | 1 | 1 | 1 |
| NES | 1 | 0 | 0 |
| look forward | 4 | 2 | 2 |
| look backward | 4 | 2 | 2 |

**Table 9: Model settings.**

| | CADEC | ShARe13 | ShARe14 |
|---|---|---|---|
| $d_h$ | 768 | 768 | 768 |
| $d_{Ed}$ | 20 | 20 | 20 |
| $d_{Et}$ | 20 | 20 | 20 |
| $d_c$ | 80 | 80 | 80 |
| Dropout | 0.5 | 0.5 | 0.5 |
| Learning rate(BERT) | $5e-6$ | $5e-6$ | $5e-6$ |
| Learning rate(other) | $1e-3$ | $1e-3$ | $1e-3$ |
| Batch size | 16 | 20 | 20 |
| warm factor | 0 | 0 | 0.1 |
| weight decay | 0 | 0.4 | 0.4 |
| epoch | 10 | 20 | 10 |
| W2NER seed | 123 | 123 | 123 |
| TOE seed | 1898 | 1898 | 1898 |